\pdfoutput=1

\documentclass[final,authoryear,12pt]{elsarticle1}

\usepackage{graphicx}
\usepackage{subfigure}
\usepackage{amssymb}
\usepackage{amsthm}
\usepackage{color}
\usepackage{booktabs} 
\usepackage{url}

\newcommand{\rev}[1]{#1}

\newcommand{\R}{\mathbb{R}}
\newcommand{\eqref}[1]{(\ref{#1})}

\journal{Pattern Recognition, Elsevier}

\begin{document}

\begin{frontmatter}

\title{Novel Classifier adapted for particular measurements.}
\author{Mat\'ias Di Martino}
\ead{matiasdm@fing.edu.uy}
\author{Guzman Hern\'andez}
\ead{guzmanhc@fing.edu.uy}
\author{Marcelo Fiori}
\ead{mfiori@fing.edu.uy}
\author{Alicia Fern\'andez}
\ead{alicia@fing.edu.uy}

\address{Facultad de Ingenier\'{i}a - Universidad de la Republica, Uruguay.}

\title{A new framework for optimal classifier design}

\begin{abstract}
The use of alternative measures to evaluate classifier performance is gaining attention, specially for imbalanced problems. However, the use of these measures in the classifier design process is still unsolved. In this work we propose a classifier designed specifically to optimize one of these alternative measures, namely, the so-called F-measure. Nevertheless, the technique is general, and it can be used to optimize other evaluation measures. An algorithm to train the novel classifier is proposed, and the numerical scheme is tested with several databases, showing the optimality and robustness of the presented classifier. 
\end{abstract}


\end{frontmatter}
\newpage

\section{Introduction}

Evaluation measures have a crucial role in classifier analysis and design.
Accuracy, Recall, Precision, F-measure, Kappa, ACU [\cite{Gar12}] and some other new proposed measures like Informedness and Markedness [\cite{Pow11}] are examples of different evaluation measures. Depending on the problem and the field of application one measure could be more suitable than another. While in the Behavioral Sciences, Specificity and Sensitivity are commonly used, in the Medical Sciences, ROC analysis is a standard for evaluation. On the other hand, in the Information Retrieval community and fraud detection, Recall, Precision and F-measure are considered appropriate measures for testing effectiveness.

In a learning design strategy, the best rule for the specific application will be the one that get the optimal performance for the chosen measure.

Looking for the best decision rule, in a Bayesian framework,
 implies to minimize the overall risk taking into
account the different misclassification cost [\cite{Dud01}];  in  an
equal misclassification cost problem we can find this optimal solution,
with maximum accuracy, selecting the class that has the maximum a
posteriori probability.

However, finding a decision rule that looks for minimum
error rate or maximum accuracy in an imbalanced domain gives solutions
strongly biased to favor the majority class, getting poor performance.

This problem is particularly important in those applications where the instances of a class (the majority one) heavily
outnumber the instances of the other (the minority) class and  it is
costly to misclassify samples from the minority class.  For example in
information retrieval [\cite{Chr08}], nontechnical losses in power
utilities [\cite{DiM12,Mun09,Nag10}] or medical diagnosis [\cite{FMAS10,Fio12b}].

 Identifying these rare events is a challenging issue with great impact regarding many problems in pattern recognition and data mining. The main difficulty in finding discriminatory rules for these applications,  is that we have to deal
with small data sets, with skewed data distributions and overlapping
classes.
A range of classifiers that work successfully
for other applications (decision trees, neural networks, support
vector machines (SVMs), etc.) get a poor performance in this context [\cite{Sun09}].  For example, in a
decision tree the pruning criterion is usually the classification
error, which can remove branches related with the minority class. In backpropagation
neural networks, the expected gradient vector length is proportional
to the class size, and so the gradient vector is dominated by the
prevalent class and consequently the weights are determined by this
class. SVMs are thought to be more robust to the class imbalance
problem since they use only a few support vectors to calculate region
boundaries. However, in a two class problem, the boundaries are determined by
the prevalent class, since the algorithm tries to find the largest margin
and the minimum error. A different approach is taken in
one-class learning, for example one class SVM, where the model is
created based on the samples of only one of the classes. In
[\cite{Ras04}] the optimality of one-class SVMs over two-class SVM
classifiers is demonstrated for some important imbalanced problems.

Recently, great effort has been done to give better solutions
to class imbalance problems (see [\cite{Sun09,Gar07,Xin08}] and references therein). In most of the approaches that deal with an imbalanced problem, the
idea is to adapt the classifiers that have good accuracy in balanced
domains. A variety of ways of doing this have been proposed: changing class distributions
[\cite{Cha02,Cha03,Kol03}], incorporating costs\footnote{The missclassification cost can be set by experts or learned \cite{Sun09}.} in
decision making [\cite{Bat04,Bar03}], and using alternative
performance metrics instead of accuracy in the learning
process with standard algorithms [\cite{Gar12}].

In this work we propose a different approach to this problem, designing a classifier
based on an optimal decision rule that maximizes a chosen evaluation measure, in this case the F-measure
[\cite{Rij79}].
\rev{More specifically, if $\Omega$ is the feature space, we are looking for the classifier $u:\Omega\to \R$ that maximizes the F-measure. Here, given the feature vector $x$, the classifier (or decision function) $u$ assigns the class $\omega_+$ if $u(x)>0$, and class $\omega_-$ if $u(x)<0$. We address this problem by proposing an energy $E[u]$ such that its minimum is achieved for the optimal classifier $u$ (in the sense of the F-measure). We solve this optimization problem using a gradient descent flow, inspired by the level-set method [\cite{Osh88}]. }
Although the analysis is made for F-measure, it could be extended to other measures. In the particular case when the chosen measure is the accuracy the proposed algorithm is equivalent to the Bayes approach.

We also show that, in contrast with common solutions, the proposed algorithm does not need to change original
distributions or arbitrarily assign misclassification costs to find an
appropriate decision rule for severe imbalanced problems. Although there is consensus about the need of using suitable evaluation measures for classifier design, to the best of our knowledge no technique has been proposed that optimizes these alternative measures over all decision frontiers.

The rest of the paper is organized as follows. In Section 2 the optimal classifier for the F-measure is proposed, and a numerical scheme to obtain it is presented. Experimental results are shown in Section 3, and we conclude in Section 4.

\section{Proposed Classifier Formulation}
In this paper we assume that there are two classes, one called here
the \textbf{negative} class, that represents the majority class,
usually associated to the normal scenario, and the other called the
\textbf{positive} class that represents the minority class. We define
$C = \{ \omega_+ , \omega_- \}$ as the set of possible classes,
being TP (true positive) the number of $x \in \omega_+$ correctly
classified, TN (true negative) the number of $x \in \omega_-$
correctly classified, FP (false positive) and FN (false negative) the
number of $x \in \omega_-$ and $x \in \omega_+$ misclassified
respectively.  Let us also recall some related well know definitions:

\begin{eqnarray*}
  \label{eq:1}
  \mbox{Accuracy:} & \mathcal{A} = \frac{TP+TN}{TP+TN+FP+FN} \label{ec:Acc} \\
  \mbox{Recall:} & \mathcal{R} = \frac{TP}{TP+FN} \label{ec:R}\\
  \mbox{Precision:} & \mathcal{P} = \frac{TP}{TP+FP} \label{ec:P}\\
  \mbox{F-measure:} & F_\beta = \frac{(1+\beta^2)\mathcal{R}\,\mathcal{P}}{\beta^2\mathcal{P}+\mathcal{R}} \label{ec:Fv}
\end{eqnarray*}
Precision and Recall are two important measures to evaluate the performance of a given classifier in an imbalance scenario. The Recall indicates the True Positive Rate, while
the Precision indicates the Positive Predictive Value. The F-measure combines them with a parameter $\beta\in[0,+\infty)$.
With $\beta=1$, $F_\beta$ is the harmonic mean between Recall and Precision, meanwhile with $\beta\gg1$ or $\beta\ll1$, the $F_\beta$ approaches the Recall or
  the Precision respectively. A high value of $F_\beta$ ensures that both
Recall and Precision are reasonably high, which is a desirable property since it indicates reasonable values of both true positive and false positive rates. \rev{The best $\beta$ value for a specific application depends on which is the adequate relation between Recall and Precision for each particular problem \citep{Chr08}.} \\
\indent The task of finding a classifier consists in defining the regions $\Omega_+$ and $\Omega_-$ of $\Omega$, such that if $x$ belongs to $\Omega_+$/$\Omega_-$, it will be classified as belonging to the positive/negative class. To train the classifier to maximize a given performance measure, we must therefore find the regions $\Omega_+$ and $\Omega_-$ that give maximal performance measure for the available data set.\\
\indent In order to find the classifier that maximizes a given performance measure, we must be able to express the quantities $FN$, $FP$ and $TP$ in terms of $\Omega_+$ and $\Omega_-$. These can be calculated by computing which points of the training data set belong to the regions $\Omega_+$ and $\Omega_-$. However, for the realization of the proposed algorithm, we will estimate these quantities in terms of probability densities for the positive and negative classes. To this end, we suppose that we have estimates for certain density functions, $f_+(x)$ and $f_-(x)$, such that in terms of these functions, we have the following approximations for the quantities $FN$, $FP$,$TP$ and $TN$:

\begin{eqnarray}
FN &=& P\int_{\Omega_-}f_+(x)dx\\
FP &=& N\int_{\Omega_+}f_-(x)dx\\
TP &=& P\int_{\Omega_+}f_+(x)dx\\
TN &=& N\int_{\Omega_-}f_-(x)dx
\end{eqnarray}

\noindent where $P$ and $N$ are the number of positive and negative instances in the training database, and the distribution functions $f_+(x)$ and $f_-(x)$ satisfy

\begin{equation}
\int_{\Omega}f_\pm(x)dx =  1
\end{equation}
If these functions are known, the task of finding the optimal classifier consists in finding the regions $\Omega_+$ and $\Omega_-$ that maximize the chosen measure. As was mentioned before, this choice depends on the particular problem or application considered. In this paper we have chosen F-measure as the evaluation measure, and in the next subsection we present an algorithm to determine the optimal boundaries for this measure. However, the framework is general, and the generalization to other evaluation measures that combine FN,FP,TN and TP is straightforward.

\subsection{Optimal boundary determination for F-measure}

\indent It can be seen that maximizing F-measure is equivalent to minimizing the quantity:

\begin{equation}
\epsilon = \frac{\beta^2FN + FP}{TP}.
\end{equation}

\noindent The quantities $FN$, $FP$, and $TP$ can be expressed in terms  of the functions $f_\pm(x)$, as was defined in the previous section. Therefore the task of training a classifier that maximizes F-measure (and minimizes $\epsilon$) can be approached as finding the regions $\Omega_+$ and $\Omega_-$ that minimize

\begin{equation}
E = \frac{k\int_{\Omega_-}f_+(x)dx + \int_{\Omega_+}f_-(x)dx}{\int_{\Omega_+}f_+(x)dx},
\label{ec:Def_E_enfunc_f1f1}
\end{equation}

where

\begin{equation}
k = \beta^2\frac{P}{N}.
\end{equation}

\indent The extent to which the quantity $E$ given by \eqref{ec:Def_E_enfunc_f1f1} is representative of the quantity $\epsilon$ depends on the extent to which the densities available, given by the functions $f_\pm(x)$ defined in the previous section, represent the distribution of points in the training data. We will not focus in this work on the task of finding appropriate probability densities, and for the sake of this paper we suppose that they are indeed available so that the quantity $E$ is a good approximation of the quantity $\epsilon$ calculated directly from the available data set.\\
\indent To perform the minimization of the quantity $E$, we express the problem in terms of an auxiliary function $u(x)$, defined so that $u(x) > 0$ if $x \in \Omega_+$ and $u(x)<0$ if $x \in \Omega_-$. For instance, the signed distance to the boundary between $\Omega_+$ and $\Omega_-$ is commonly used in the implementation, since it has proven to give good results. The boundary between the regions $\Omega_+$ and $\Omega_-$ is therefore given by the surface which satisfies the equation $u(x) = 0$. Definition \eqref{ec:Def_E_enfunc_f1f1} may be thus expressed as a functional of $u(x)$,

\begin{equation}
E[u] = \frac{k\int H_\epsilon(-u(x))f_+(x)dx + \int H_\epsilon(u(x))f_-(x)dx}{\int H_\epsilon(u(x))f_+(x)dx},
\end{equation}

\noindent where $H_\epsilon(y)$ is a smoothed Heavyside function and the domains of integration are now all $\Omega$. In these terms, the task of training the classifier consists in finding a function $u_m(x)$ which minimizes this functional. To this end, we must find the function $u_m(x)$ that cancels the first variation of the functional $E[u]$, which can be written in terms of the functional derivative of $E[u]$. Calculating this functional derivative we have:

\begin{equation}
E'[u(x)] = \frac{1}{\int H_\epsilon(u)f_+(x)dx}\delta_\epsilon(u(x))[f_-(x) - (k + E[u])f_+(x)]
\end{equation}

\noindent where $\delta_\epsilon(y)$ is the derivative of $H_\epsilon(y)$, that is, a smoothed Dirac delta function. To solve the minimization problem, we must now find the classifier function $u_m(x)$ that satisfies

\begin{equation}
\label{euler}
E'[u_m(x)] = 0
\end{equation}

\subsection{Implementation}

The classical gradient descent flow method is used in order to solve the Euler-Lagrange equation \eqref{euler}. Specifically, the following PDE (Partial Differential Equation) is solved with a certain initialization $u_0$ :

\begin{equation}
\label{flujo}
\left \{
\begin{array}{l}
\frac{\partial u(x,t)}{\partial t} = -E'[u(x,t)] \\
u(x,0) = u_0(x)
\end{array}
\right.
\end{equation}

When the steady estate of this PDE is reached, equation \eqref{euler} is satisfied (see [\cite{Sap01}] for more details). 
\rev{Since equation \eqref{euler} is to be solved numerically, in principle any sufficiently regular densities $f(x)$ are allowed, and therefore the proposed algorithm does not depend on the particulars of the density estimation process.}

The introduction of the auxiliary function $u(x)$ is motivated by the Level Set Method [\cite{Osh88}], and although it is not the same kind of curve evolution, these approaches share some known implementation details that must be taken into account. For instance, in order to guarantee stability, it is usual to reinitialize (after a certain amount of iterations) the function $u(x)$ in order to keep it as a distance function. \rev{The only relevant information of $u(x)$, in terms of the evaluation of the functional $E[u]$, is the partition $(\Omega_+,\Omega_-)$ that $u(x)$ defines. Therefore, it is possible to reinitialize the function $u(x)$ to the signed distance function to the zero-level set, since this keeps the sign of $u(x)$ unchanged, and therefore the classifier and the energy $E[u]$ remain unchanged.}  
For the explicit scheme and more details see [\cite{Sus94}].

Another usual practice is to add a regularization term $\Delta u$ to the flow \eqref{flujo} (corresponding to a Tikhonov term in the functional [\cite{Tik77}]). This latter is a minor detail that does not significantly affect the resulting function $u$.

The resulting numerical scheme to solve \eqref{euler} is then:

$$
u^{n+1}= u^n - \Delta_t G
$$
where
$$
G = \delta_\epsilon(u^n)(f_2-\beta^2 f_1)\int_\Omega{f_1H_\epsilon(u^n)dx - \delta_\epsilon(u^n)f_1\int_\Omega{[f_2H_\epsilon(u^n)+\beta^2f_1H_\epsilon(-u^n)]dx + \lambda\Delta u}}
$$

and $\Delta_t$ is the time step. This iterative algorithm is repeated until convergence (i.e. the difference between $u^n$ and $u^{n+1}$ is small).

\begin{figure}[h!]
\centering
\subfigure[Initialization]{
\centering \includegraphics[width=.3\textwidth]{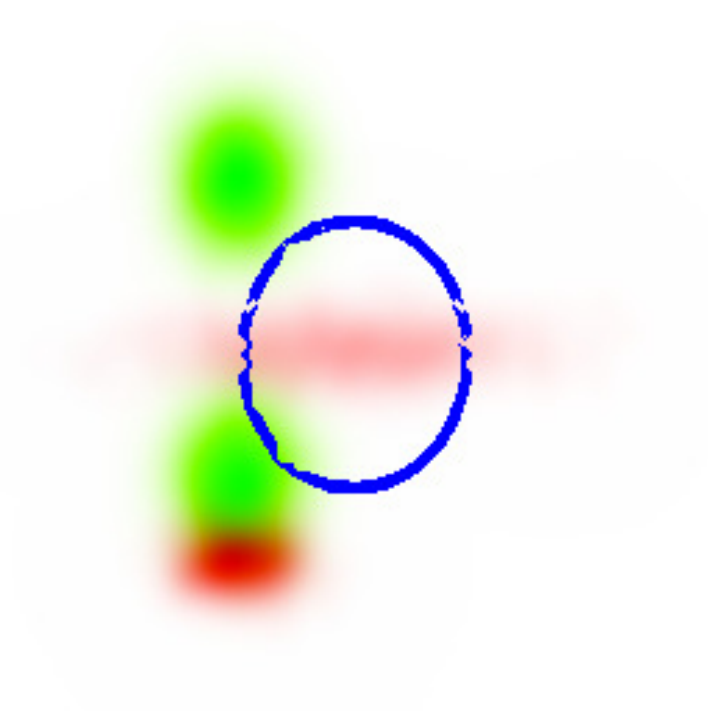}
}
\subfigure[After $100$ iterations.]{
\centering \includegraphics[width=.3\textwidth]{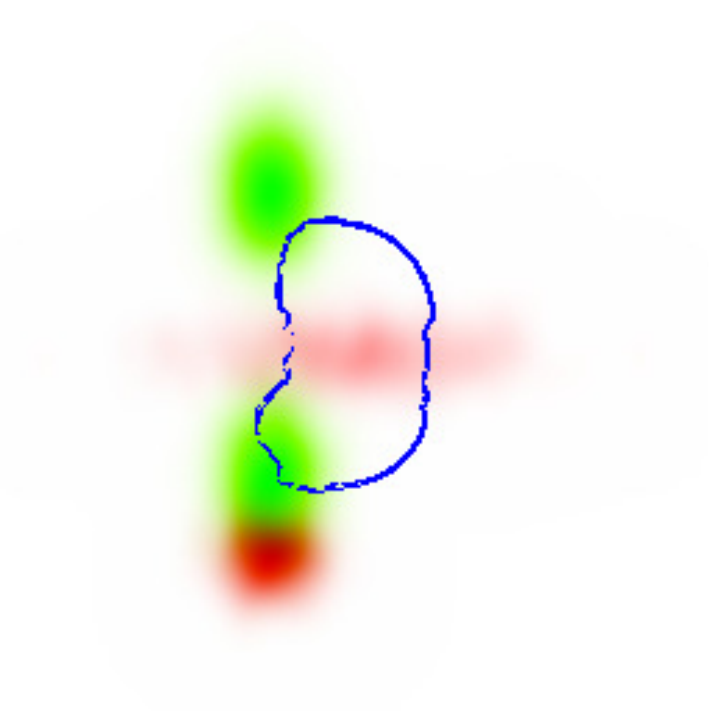}
}
\subfigure[After $600$ iterations.]{
\centering \includegraphics[width=.3\textwidth]{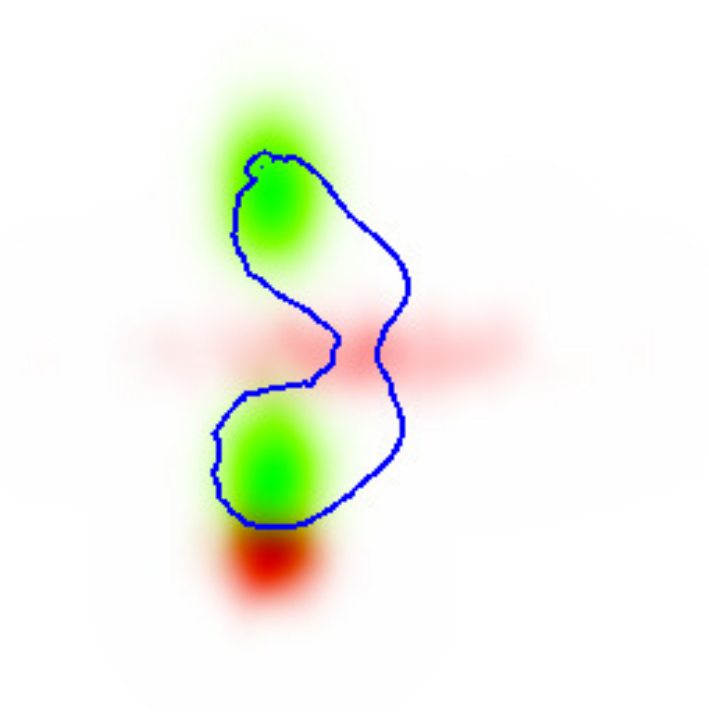}
}
\subfigure[After $700$ iterations.]{
\centering \includegraphics[width=.3\textwidth]{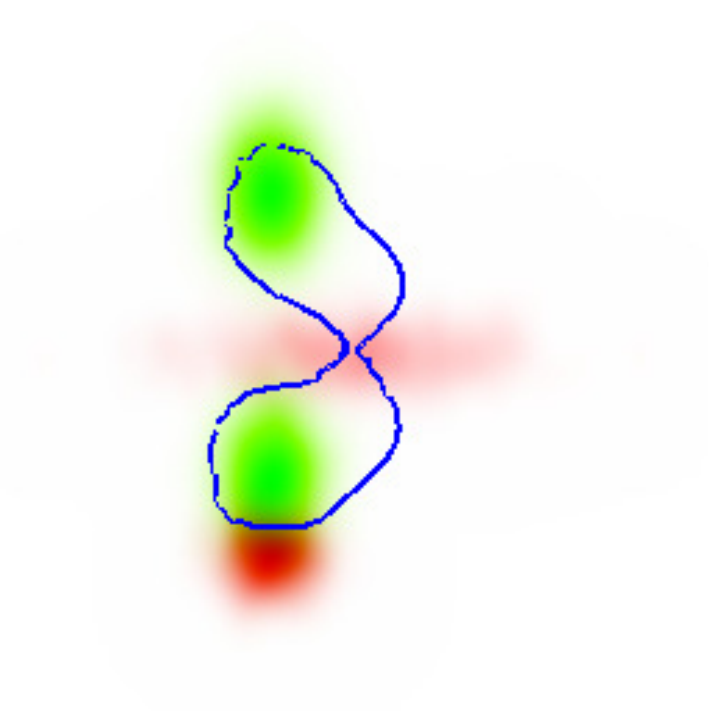}
}
\subfigure[After $800$ iterations.]{
\centering \includegraphics[width=.3\textwidth]{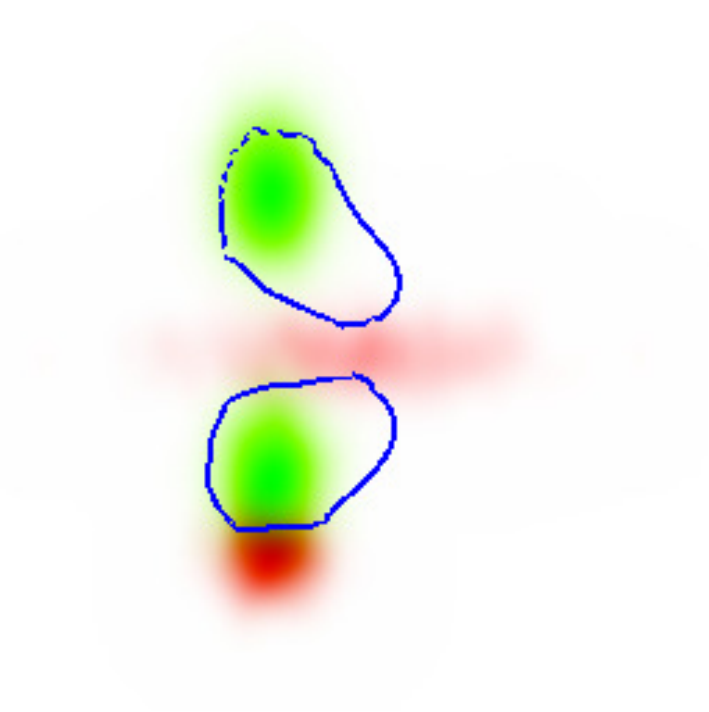}
}
\subfigure[After $1200$ iterations.]{
\centering \includegraphics[width=.3\textwidth]{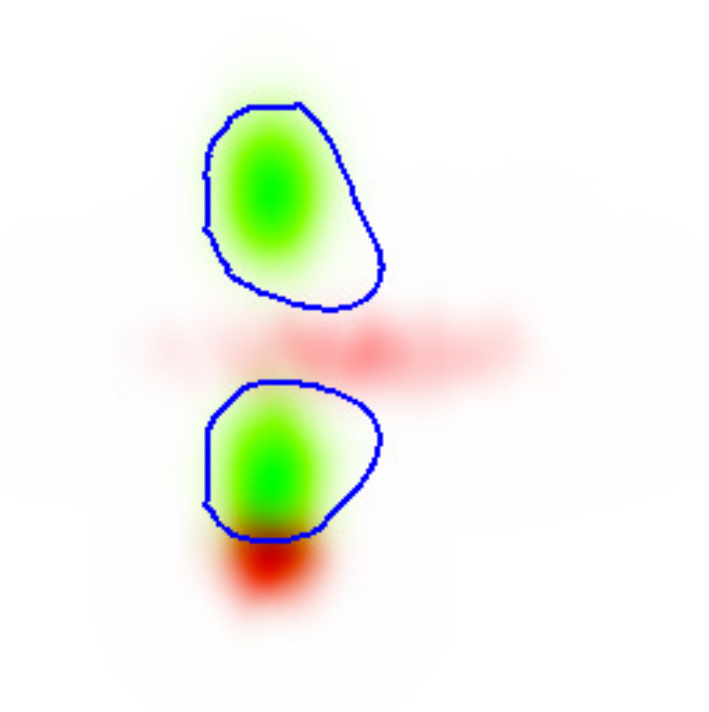}
}
 \caption{Evolution of the zero level set of $u$ (decision frontier).}\label{fig:evolucion}
\end{figure}

At each time $t$, the zero level set of $u(x,t)$ is the decision frontier of the classifier. In Figure \ref{fig:evolucion}, the evolution of this frontier is shown, from the initial $u_0$ to the final $u(x,T)$, for a certain database (described in the next section). The densities of the positive and negative classes are represented in green and red respectively.

\rev{Although we have no rigorous proof on the existence of a solution to the equation provided}, we have exhaustive empirical evidence that if the zero level set of the initialization $u_0$ includes or intersects all the connected components of the support of either one of the densities, then the gradient descent flow converges to the global optimum.

The code is available at \url{www.fing.edu.uy/~matiasdm}.

\section{Experimental Results}

\subsection{Synthetic Data}
\subsubsection{Data description}

\begin{figure}[h!]
\subfigure[Database 1]{
\centering\includegraphics[width=.22\textwidth]{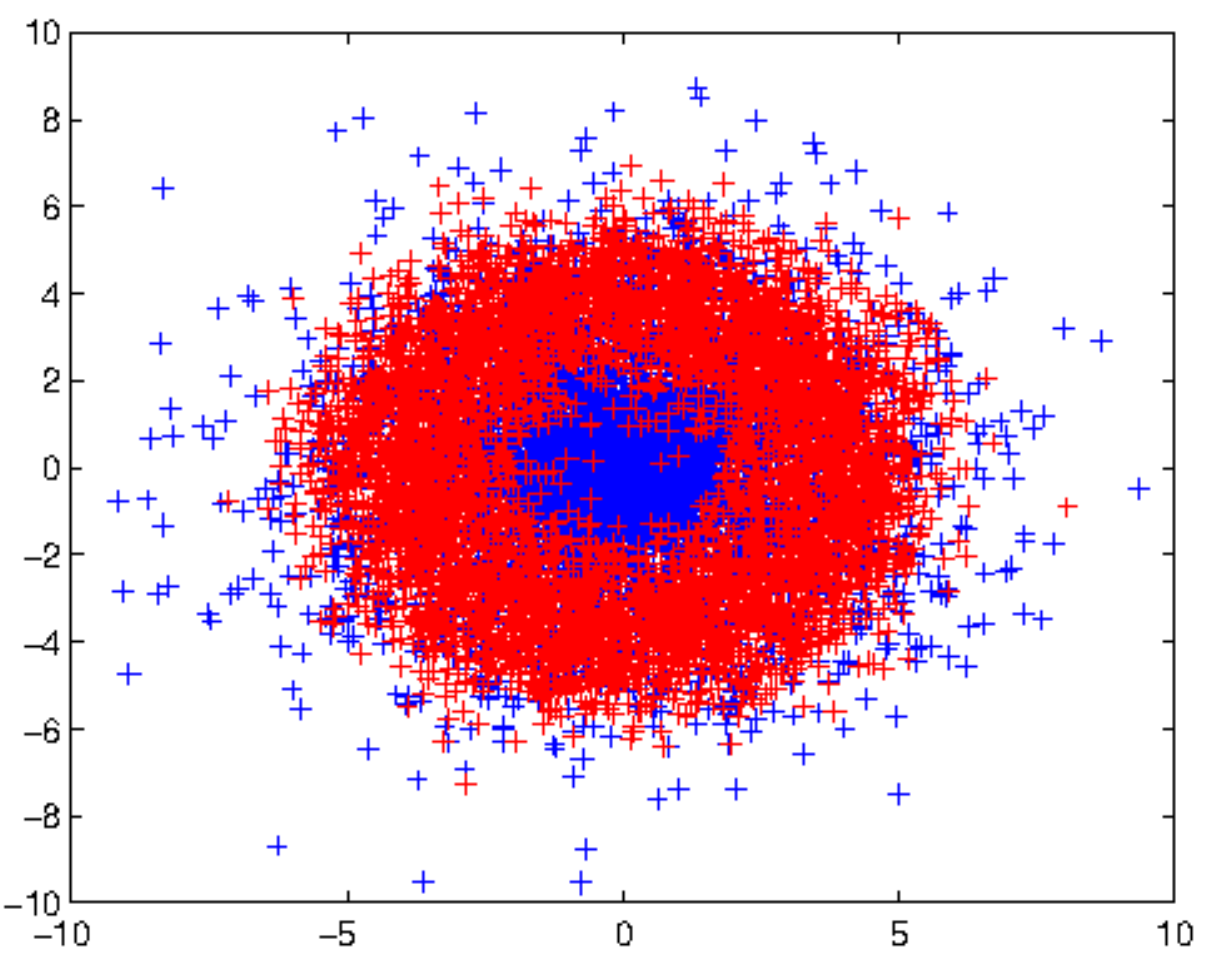}
\label{fig:base_1}
}
\subfigure[Database 2]{
\centering\includegraphics[width=.22\textwidth]{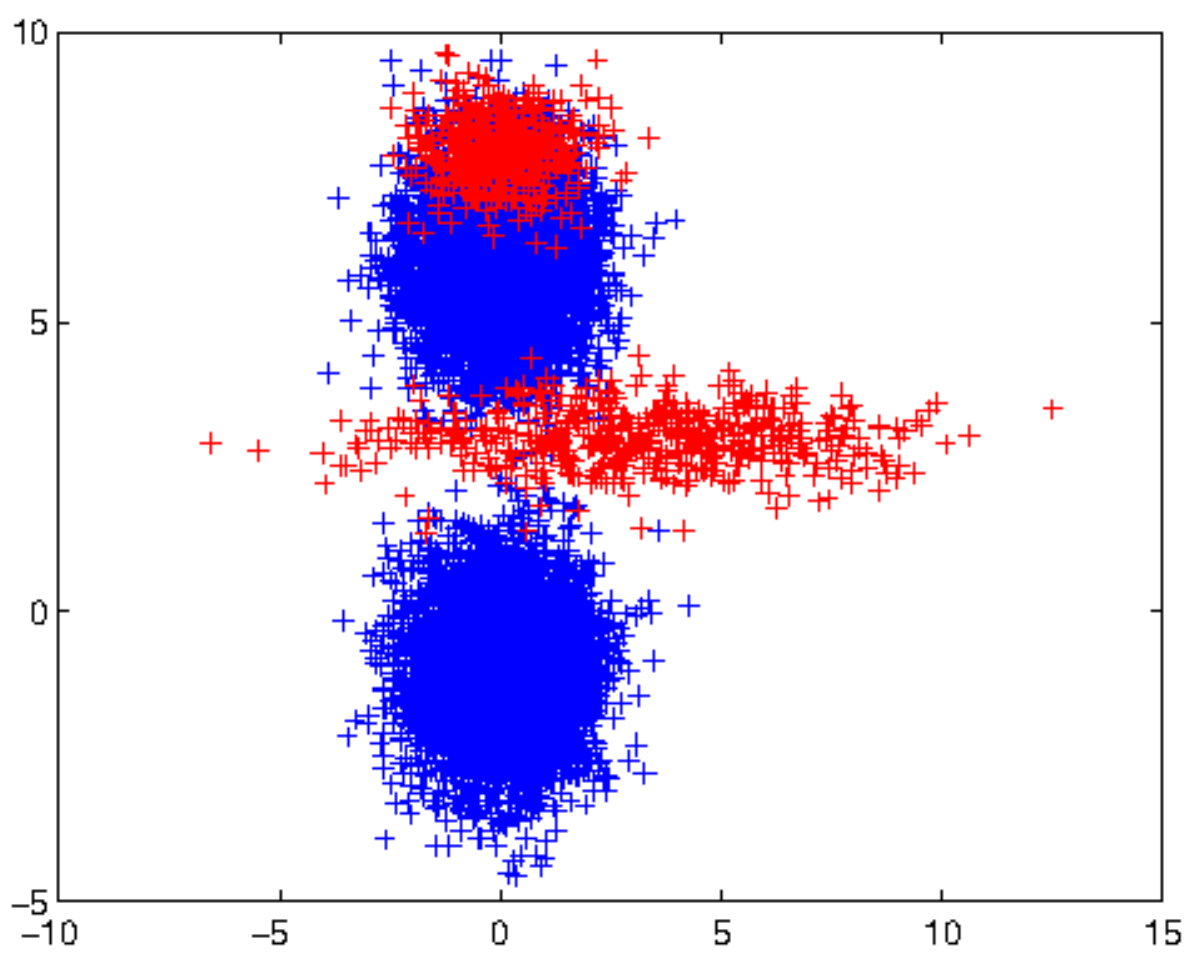}
\label{fig:base_2}
}
\subfigure[Database 3]{
\centering\includegraphics[width=.22\textwidth]{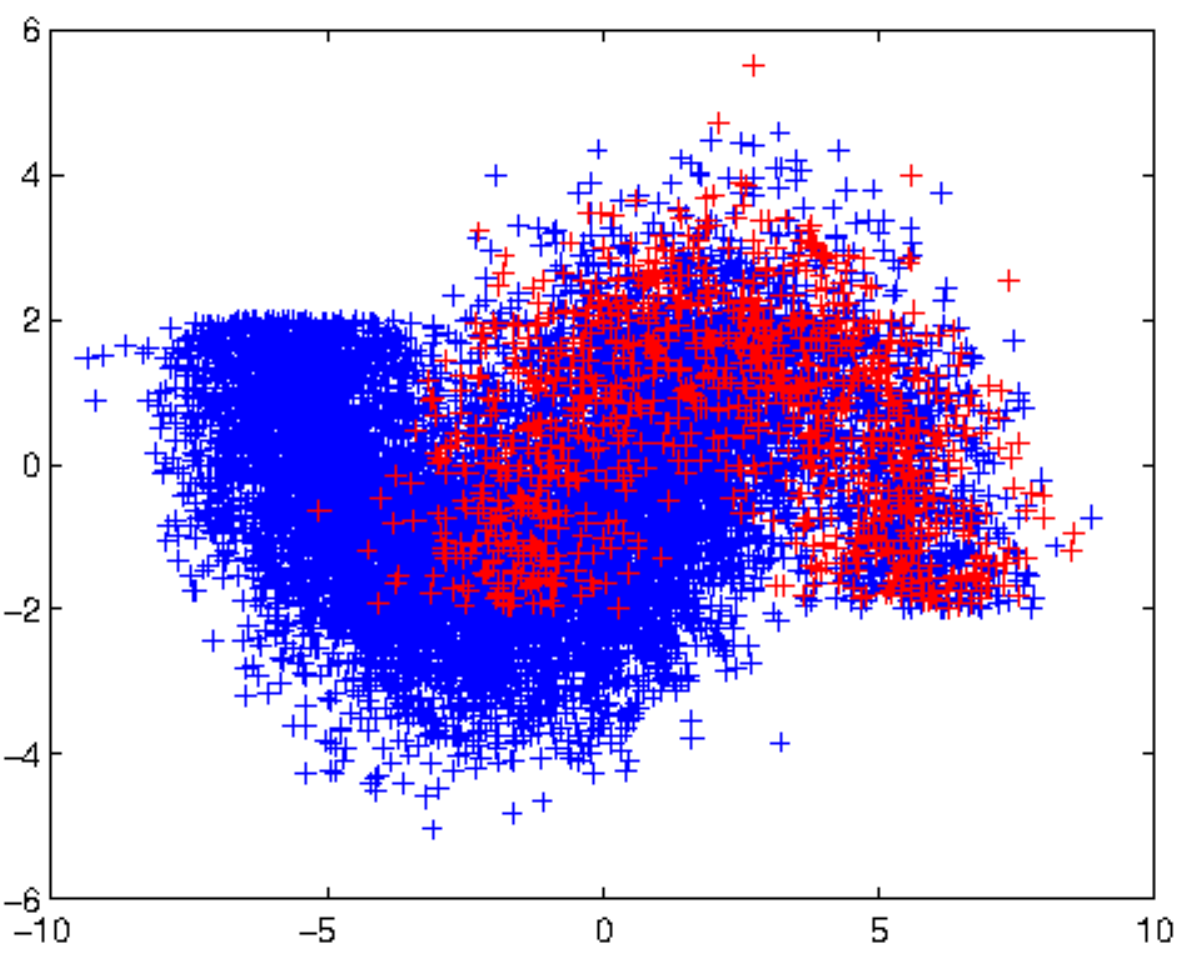}
\label{fig:base_3}
}
\subfigure[Database 4]{
\centering\includegraphics[width=.22\textwidth]{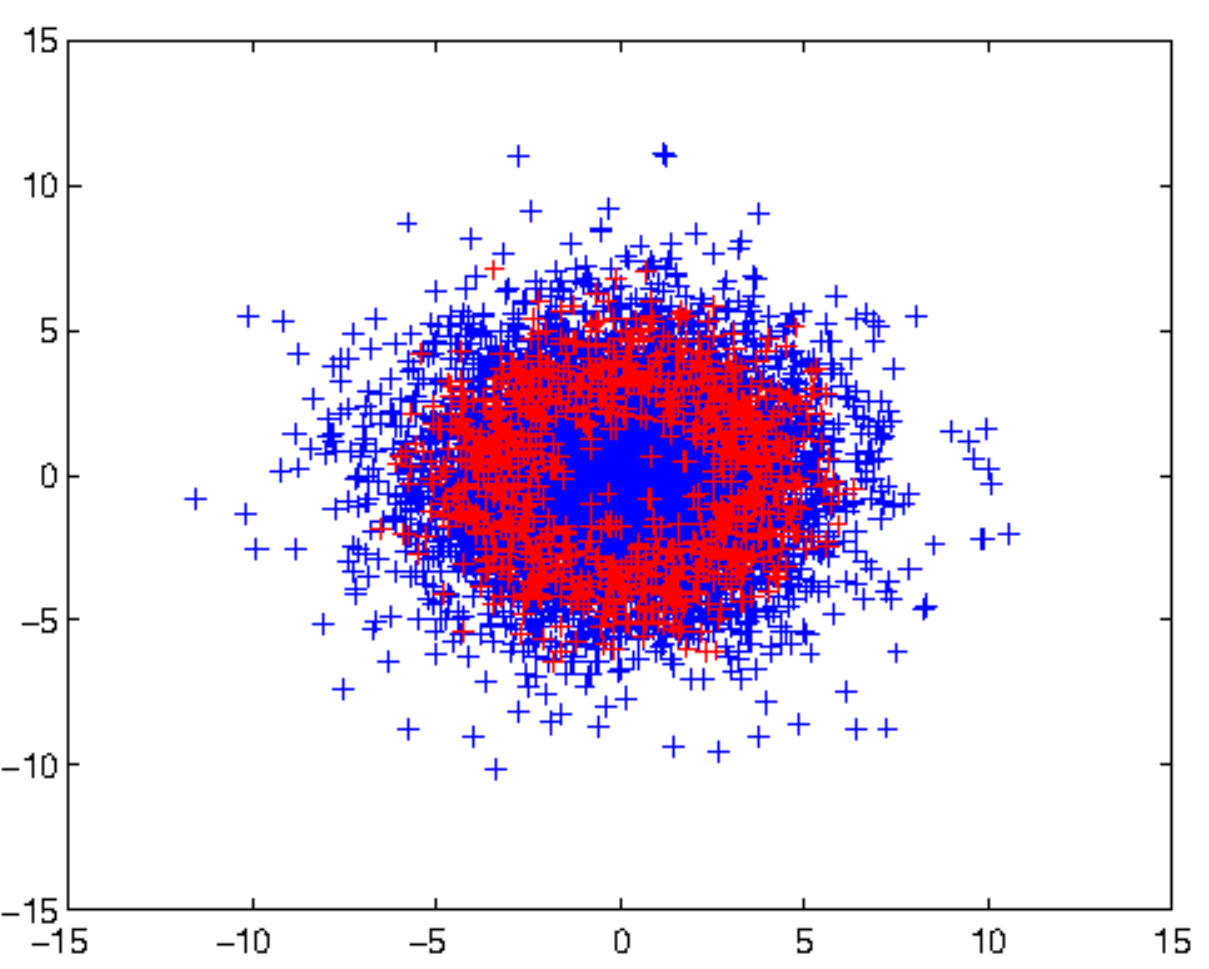}
\label{fig:base_4}
}
\caption{Databases}\label{fig:bases}
\end{figure}

For the experimental validation, we used the four different databases shown in Figure \ref{fig:bases}. Database 1 has a negative class with a Gaussian distribution while the positive samples has a \emph{ring} distribution (Figure \ref{fig:base_1}). In this particular case there are $5000$ samples of the negative class and the same amount of the positive class. Database 2 has a multimodal distribution for both the positive and negative samples. For this database there are $10000$ samples of the negative class and $1000$ samples of the positive class. The third database has a \emph{horseshoe} distribution with $10000$ samples of the majority class and $1000$ samples of the minority class. The last database has the same distributions as database 1, but with $10000$ negative samples and $1000$ positive samples.\\
The selected databases do not play any particular role, the idea was to consider different scenarios such as: imbalance (Databases 2-4) and \emph{balance} (Database 1), and also evaluate a wide variety of shapes for the classes distributions. In these experimental comparisons, a classical kernel density estimation technique was used to infer the densities of the positive and negative classes \citep{Wan94}.

\subsubsection{Numerical results}
We compare the proposed algorithm, from now on called OFC (acronym for Optimal F-measure Classification), with One Class SVM (with and without kernel), the C45 tree and the traditional Naive Bayes classifier. The parameters for each algorithm were chosen to maximize the F-measure (performing 10-fold cross validation). In the next subsection we will briefly explain why we chose those algorithms and what considerations must be taken into account before the performance comparison.

Table \ref{tb:resultados_base_anillo} shows in detail the results obtain for the Database 4. Each algorithm was run 10 times (and for each execution 10-fold cross validation was performed), using $\beta=1$ and $\varepsilon=10^{-5}$. As it can be seen, the best F-measure was obtained for OFC (as expected) followed by the One Class SVM classifier. 

\begin{table}[h!]
\centering\begin{tabular}{|c|c|c|c|c|}
  \hline
   Classifier: & $F_{\beta}$ &  Acc & Rec & Pre \\
  \hline
  OFC & $\mathbf{33.67\pm0.14}$ & $71.98\pm0.10$ & $78.25\pm0.44$ & $21.45\pm0.09$ \\
  \hline
  C45 & $18.64\pm0.79$ & $87.89\pm0.17$ & $15.26\pm0.73$ & $\mathbf{23.98\pm0.99}$ \\
  \hline
  OSVM & $25.30\pm0.50$ & $62.13\pm0.62$ & $70.58\pm2.41$ & $15.41\pm0.27$ \\
  \hline
  OSVM+ker & $31.97\pm0.68$ & $67.17\pm1.63$ & $\mathbf{84.76\pm1.77}$ & $19.71\pm0.60$\\
  \hline
  N. Bayes & $1.54\pm0.42$ & $\mathbf{90.61\pm0.02}$ & $0.81\pm0.23$ & $16.37\pm2.99$ \\
  \hline
 \end{tabular}\caption{Performance values ($\%$) over 10 executions of each algorithm using database 4. $\beta =1$}\label{tb:resultados_base_anillo}
\end{table}

\begin{figure}[h!]
\centering
  \def\svgwidth{250pt}
  \begin{small}
  \input{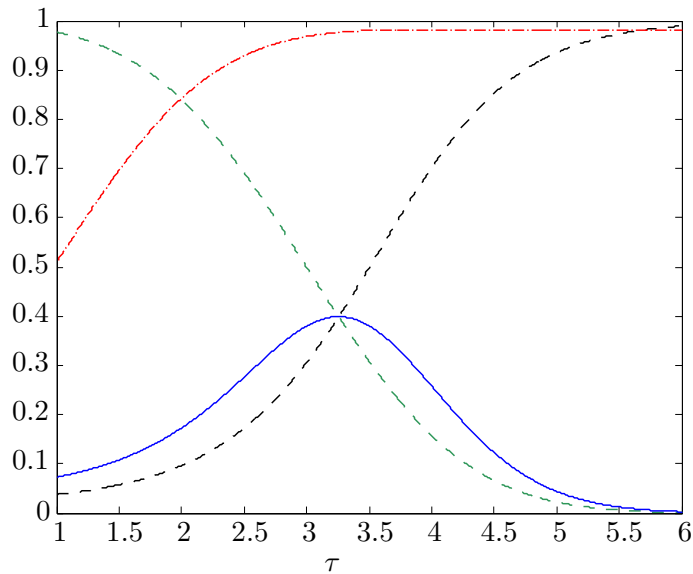}
  \end{small}
  \caption{Performance measures for several values of the decision threshold $\tau$, for the unidimensional problem with $\beta=1$. F-measure in blue (solid), Recall in green (decreasing dashed), Precision in black (increasing dashed) and Accuracy in red (dash-dot).}
  \label{toy_example}
\end{figure}

It is worth mentioning the Naive Bayes performance. It is the algorithm with the best accuracy, which is expectable, but with the poorest F-measure. This is the typical behavior of those classifiers which are designed for minimizing the classification error in problems were the classes are highly overlapped and unbalanced.
{
\rev{
To illustrate this point we consider a 1-D problem with Gaussian distributions for both the negative and positive classes, with means 1 and 3 respectively, and the unitary variance. The number of samples is 1000 for the positive class and 50000 for the negative class. The decision problem (i.e. the determination of the regions $\Omega_+$ and $\Omega_-$) in this toy example amounts to choosing a decision threshold $\tau$ which sets the frontier between the classes in the real line, so that $\Omega_- = \{-\infty,\tau\}$ and $\Omega_+ = \{\tau,\infty\}$. So for different values of $\tau$, one would get different values of the Accuracy, Recall, Precision and F-measure. Figure \ref{toy_example} shows these dependencies as a function of this decision threshold. We can see that the OFC solution is the one that gives the best F-measure, with a good tradeoff between recall and precision (consistent with the $\beta=1$ chosen) and a loss of approximately $0.5\%$ of Accuracy compared with the optimal accuracy that could 
be obtained by the Naive Bayes solution ($\tau=3.96$). Getting a better Accuracy or Precision, but very bad recall, could be a bad solution when the positive class is the relevant one (cancer lesion, fraud samples). We can also see from this figure that setting the threshold away from the optimal F-measure point it is possible to get a better value of Precision, sacrificing the value of the Recall, and conversely. This is consistent with the result found for OSVM+ker shown in Table \ref{tb:resultados_base_anillo}, which has slightly lower F-measure than OFC, getting in this way a higher Recall yet lower Precision.}
}

\begin{figure}[h!]
\centering\includegraphics[width=.8\textwidth]{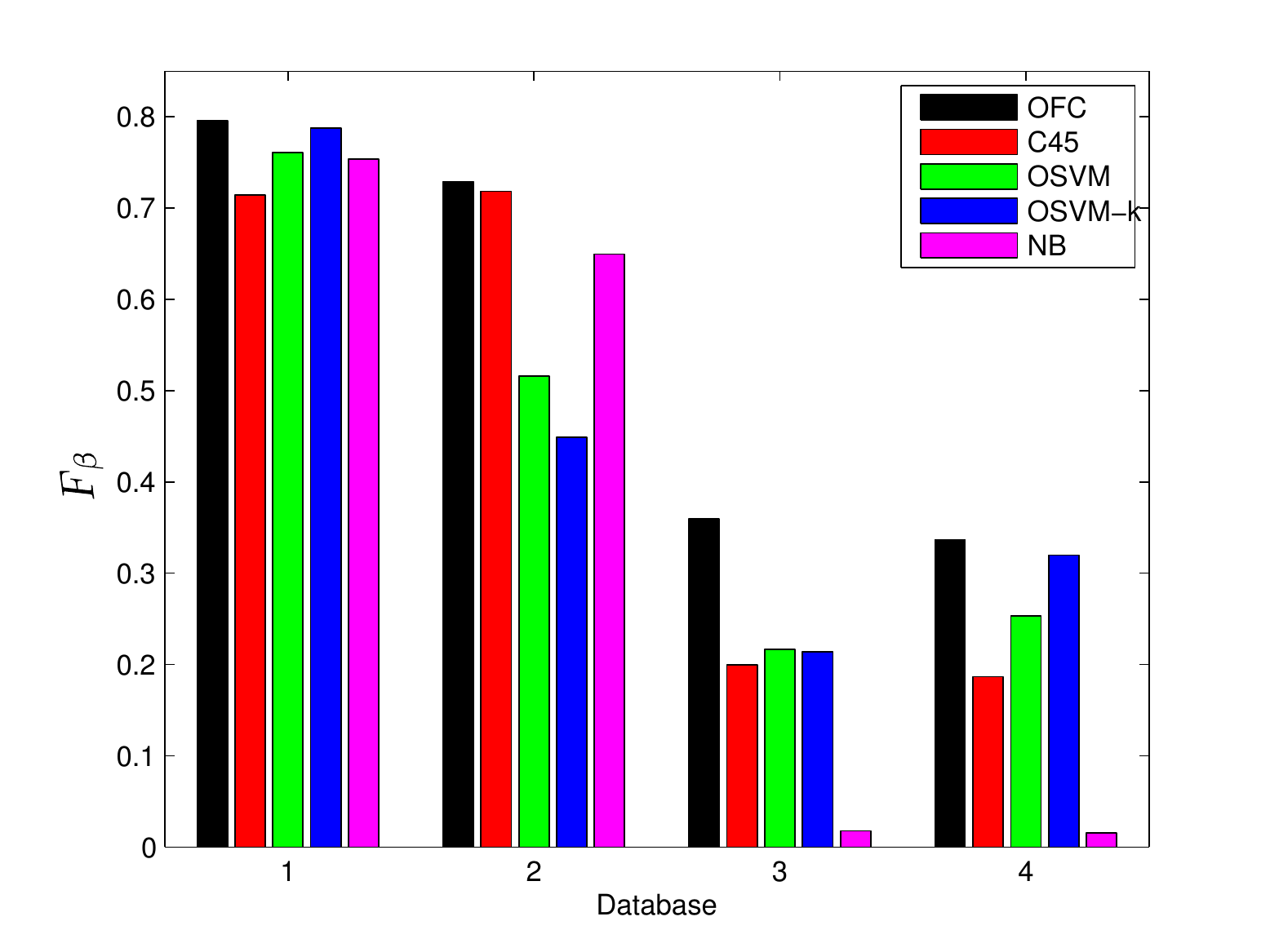}
\caption{$F_\beta$ values for different classifiers using databases 1-4} \label{fig:comparacion}
\end{figure}

In Figure \ref{fig:comparacion} the mean $F_\beta$ values obtained over 10 executions using databases 1-4 are shown. The standard deviations were under $1\%$ in all cases. As was explained above, when the classes have similar amounts of samples, or separable distributions (databases 1-2), the differences between the \emph{traditional algorithms} (C4.5 - NB) and those designed for imbalance problems (OSVM - OFC) is not so important, while in the other cases (databases 3-4) the difference became more significant.

\begin{figure}[h!]
  \def\svgwidth{350pt}
  \input{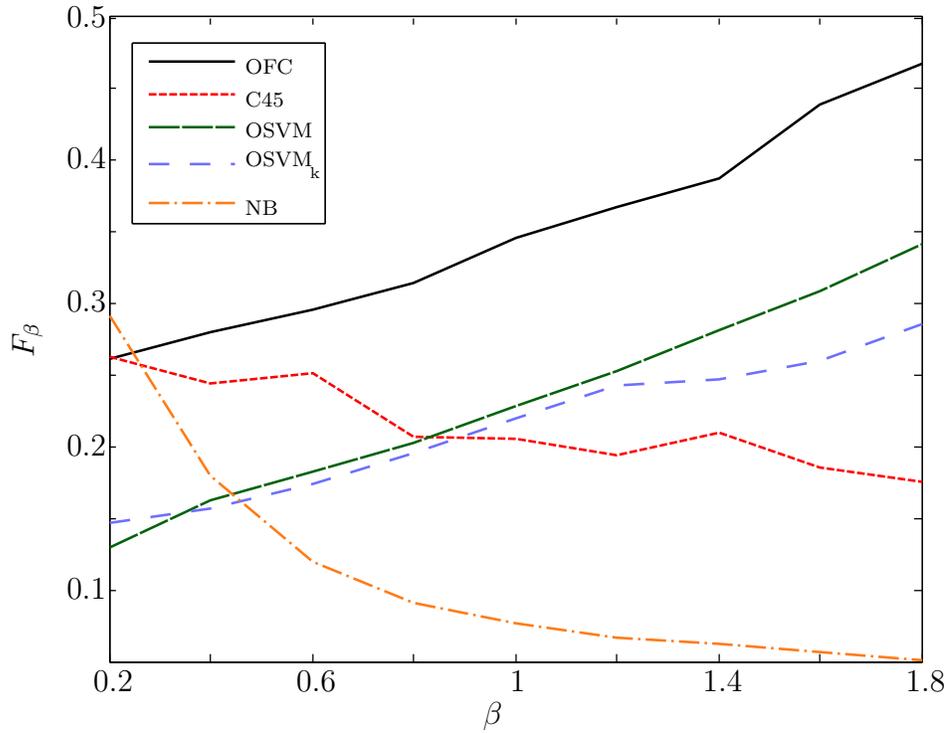}
  \caption{$F_\beta$ performance of the different classifiers, for several values of $\beta$.}
  \label{fig:varios_betas}
\end{figure}

Finally the Figure \ref{fig:varios_betas} shows an additional experiment that illustrates the robustness of the algorithm when varying $\beta$ (which changes the weight of the Recall and Precision in the $F_\beta$ definition). For this experiment, database 3 was used.

{
\rev{
\subsection{Experiment with skin segmentation data}
To conclude this section, we present an additional experiment with
skin segmentation data [\cite{Raj10}] from the UCI Machine Learning
Repository. The skin dataset was collected by randomly sampling R,G,B (red, green, blue)
values from face images of various age groups (young, middle, and
old), race groups (white, black, and asian), and genders obtained from
FERET database and PAL database. Total sample size is 245057 samples;
out of which 50859 correspond to skin samples and 194198 to non-skin samples.
The results are shown in Figure \ref{fig:result_skin}, where OFC and OSVM are compared for several values of $\beta$. One class SVM\footnote{As in the previous experiments OSVM parameters were
  set using cross validation selecting those parameters that gives the
  highest F-measure} achieves the highest Recall but with a poor Precision, therefore obtaining a low F-measure, while our approach outperforms OSVM in terms of F-measure as expected. 
  Observe that for values $\beta \gg 1$, maximizing the F-measure is equivalent than maximizing the Recall, and therefore both approaches (OSVM and OFC) are practically equivalent. In
addition, the time required for OFC was approximately ten times lower
than for OSVM.
\begin{figure}[h!]
 \centering\includegraphics[width=.75\textwidth]{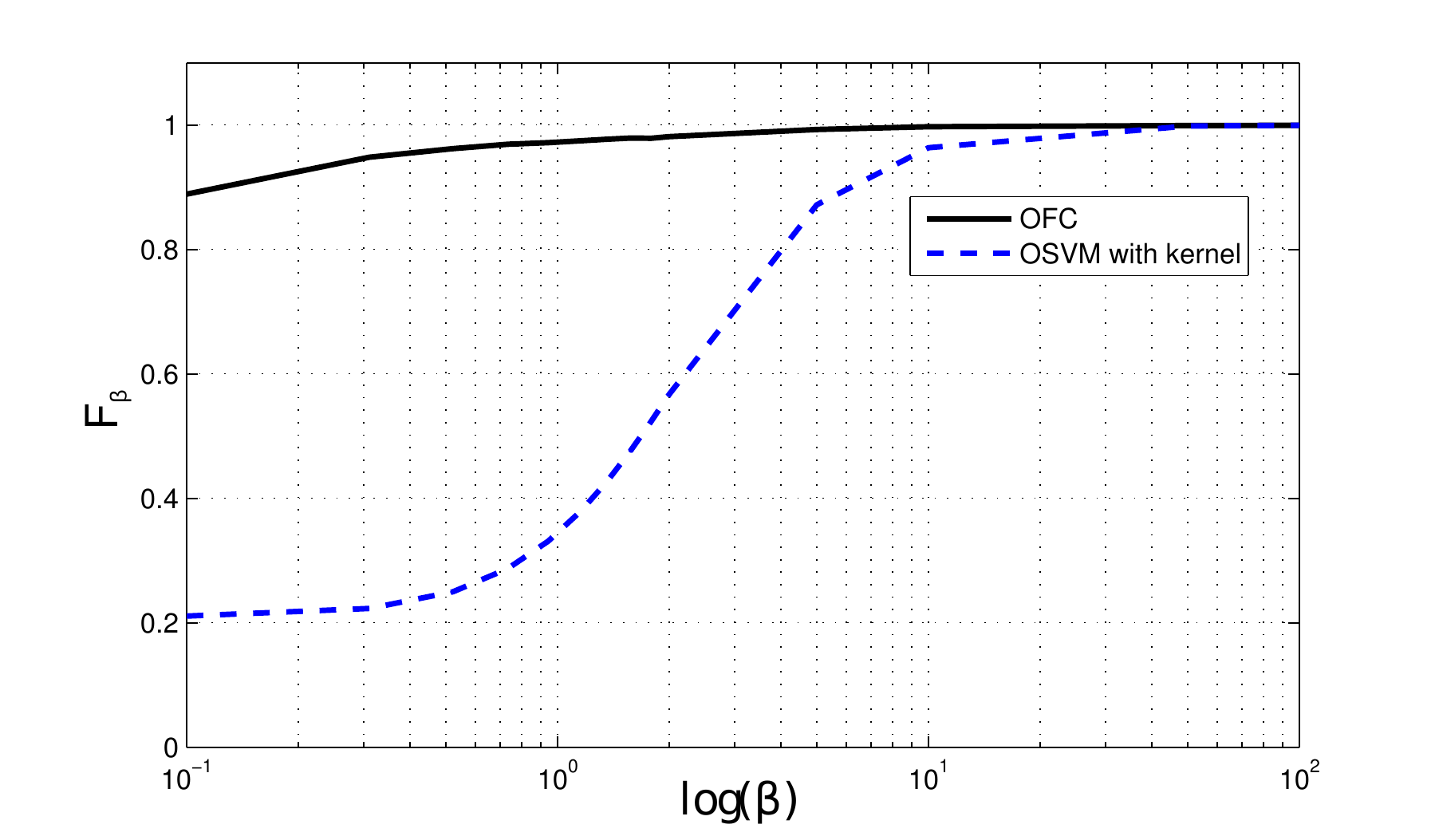}
 \caption{Algorithm comparison using skin segmentation data. F-measure
   for OFC (in black) and One Class SVM (in blue)}\label{fig:result_skin}
\end{figure}
}
}
\subsection{Analysis and considerations}
In the previous subsection the results for different databases were provided, showing that the proposed algorithm is suitable for imbalanced problems.
Even though in this work we include the results obtained for the algorithms C45, Naive Bayes and One Class SVM (with and without kernel) for the sake of completeness, we consider that the performance comparison should be done with One Class SVM, since the other algorithms are not designed for imbalanced problems.

The results of Naive Bayes and C45 reinforce the well-know behavior: \emph{traditional approaches} have good performance in the most common (balanced) problems, but they are not adequate for imbalanced problems.

On the other hand, several techniques are proposed in the literature to improve the performance of this type of algorithms in unbalanced scenarios, such as SMOTE, ADABOOST, SMOOTEbost among others (see \cite{Cha02,Cha03,Ham07,Lop12,Xin08,Gar06,Gar07,Gar12} and references therein for more details). However, all these methods are pre or post-processing techniques that use the base classifiers as black boxes, and the main point of this section is to compare these base classifiers by themselves.

\rev{
In terms of the computational performance of the algorithm proposed, through the examples studied it was found that the algorithm (as implemented for the tests realized) runs very efficiently in low dimensions, for instance running much faster than the OSVM algorithm used to compare performances in the example using skin segmentation data. However, it must be noted that the memory storage of our implementation depends on the size of the grid used to compute the decision function $u(x)$. Nevertheless, efficient solutions to this problem are available, for instance allowing to evaluate the kernel density estimation at $m$ evaluation points from $n$ sample points in $O(n+m)$ \cite{Ray10}}.

\section{Conclusions and Future Work}

We have proposed a new framework for classification in imbalanced problems, and classifier design in general. We presented the optimality conditions for the decision frontier to maximize the F-measure, and a numerical scheme to solve the problem.

The technique is general, in the sense that it can be used to obtain optimal classifiers with respect to other evaluation measures (in addition to the F-measure).

The analysis is supported by experimental results, which show the potential and practical use of the proposed scheme.

There are other important properties and experiments to consider, making it interesting to further study the proposed framework. For instance, the feasibility and convenience of using kernels with the proposed classifier is subject of future research, as well as the combination of the proposed framework with other techniques used to improve traditional classifiers (such as SMOTEboost or ADABOOST).

The application of the optimal $F_{\beta}$ classifier to other very important problems, such as fraud detection (\cite{DiM12}) and polyp detection \cite{Fio12} is part of future work as well.




\newpage

\bibliographystyle{elsarticle-harv}
\bibliography{bibfile}

\end{document}